\documentclass{article}

\usepackage{amsmath,graphicx,mlspconf}
\usepackage{cite}
\usepackage{amssymb,amsfonts}
\usepackage{algorithmic}
\usepackage{textcomp}
\usepackage{xcolor}
\usepackage{hyperref}
\usepackage{subfig}
\usepackage{adjustbox}

\definecolor{lavendar}{HTML}{8E7DBE}
\definecolor{pine}{HTML}{2F3E46}
\definecolor{grassgreen}{HTML}{4B7F52}
\definecolor{peach}{HTML}{FAAA8D}
\newsavebox{\measurebox}
\toappear{2024 IEEE International Workshop on Machine Learning for Signal Processing, Sept.\ 22--25, 2024, London, UK}

\title{Modeling Human Responses by Ordinal Archetypal Analysis}
\name{%
\parbox{\linewidth}{\centering
      Anna Emilie J. Wedenborg$^{1}$$^{\star}$$^{\dagger}$,
      Michael Alexander Harborg$^{1}$$^{\star}$,
      Andreas Bigom$^{1}$,
      Oliver Elmgreen$^{1}$,
      Marcus Presutti$^{1}$,
      Andreas Råskov$^{1}$,
      Fumiko Kano Glückstad$^{2}$,
      Mikkel Schmidt$^{1}$ and
      Morten Mørup$^{1}$%
    }\thanks{This work was supported by the Danish Data Science Academy, which is funded by the Novo Nordisk Foundation (NNF21SA0069429) and VILLUMFONDEN (40516).}%
}
\address{%
    $^{1}$ The Technical University of Denmark \hspace{1cm}
    $^{2}$ Copenhagen Business School \\%
    $^{\star}$ Equal work contribution
    \hspace{1cm}
    $^{\dagger}$ Corresponding author:$aejew@dtu.dk$%
}
\copyrightnotice{979-8-3503-7225-0/24/\$31.00 {\copyright}2024 IEEE}
\begin{document}

\maketitle

\begin{abstract}
This paper introduces a novel framework for Archetypal Analysis (AA) tailored to ordinal data, particularly from questionnaires. Unlike existing methods, the proposed method, Ordinal Archetypal Analysis (OAA), bypasses the two-step process of transforming ordinal data into continuous scales and operates directly on the ordinal data. We extend traditional AA methods to handle the subjective nature of questionnaire-based data, acknowledging individual differences in scale perception. We introduce the Response Bias Ordinal Archetypal Analysis (RBOAA), which learns individualized scales for each subject during optimization. The effectiveness of these methods is demonstrated on synthetic data and the European Social Survey dataset, highlighting their potential to provide deeper insights into human behavior and perception. The study underscores the importance of considering response bias in cross-national research and offers a principled approach to analyzing ordinal data through Archetypal Analysis.


\end{abstract}
\begin{keywords}
Archetypal Analysis, Ordinal data, response bias, questionnaires
\end{keywords}
\section{Introduction}

Archetypal Analysis (AA) seeks to uncover distinct aspects in data defined as convex combinations of the observations within a given dataset such that the data can be reconstructed as convex combinations of these unique features. Thereby, AA forms a latent polytope with corners defined by the extracted distinct aspects, commonly referred to as “archetypes”. They play a pivotal role in understanding the intrinsic structure of the data \cite{Cutler1993ARCHETYPALANALYSIS}, providing a continuous spectrum in terms of how each observation is characterized by the extracted features \cite{Cutler1993ARCHETYPALANALYSIS,Mrup2010ArchetypalLearningb,Mrup2012ArchetypalMining}.

AA has traditionally been applied to continuous data types, yet recent advancements have facilitated its application to a broader spectrum of data distributions. Notably, AA has been extended into a probabilistic framework that optimizes pseudo-likelihoods (i.e., likelihood functions in which the observations depend on the observations themselves), allowing for the analysis of discrete data \cite{Seth2016ProbabilisticAnalysis}. Recent expansions have been particularly significant in the realm of questionnaire data, where frameworks have been developed for both nominal \cite{Seth2016ArchetypalObservations} and ordinal data \cite{Fernandez2021ArchetypalData}. The latter method will be used as a benchmark for our results. It involves a two-step approach: initially, ordinal data is transformed into a continuous scale via an Expectation-Maximization (EM) clustering algorithm. This preliminary step necessitates the determination of an optimal cluster count, which can be a meticulous process. Subsequent to this transformation, conventional least squares AA is applied to the transformed continuous data. 

Techniques such as Principal Component Analysis (PCA) and Nonnegative Matrix Factorization (NMF) have been refined to explicitly incorporate ordinal datasets\cite{Linting2007NonlinearApplication,Gouvert2020OrdinalRecommendation}. Likewise, adaptations have been made to Gaussian Process models to support ordinal data\cite{Chu2005GaussianRegression}. However, these approaches do not consider the individual variances in scale interpretation when applied to questionnaire data that captures subjective self-assessments. Furthermore, they lack the natural interpretability characteristic of Archetypal Analysis.

We propose a direct optimization framework for AA for ordinal data, circumventing the two step procedure explored previously \cite{Fernandez2021ArchetypalData}. Our method is further distinguished by its recognition of the subjective nature of questionnaire data, acknowledging that different individuals may perceive ordinal scales differently. This perspective is particularly pertinent in the context of self-reported data, where subjects are prompted to express their opinions on a Likert-type scale. Traditional methodologies have often treated such data as if it were continuous, presupposing uniform intervals between scale points. This assumption is a simplification that overlooks the ordinal nature of the data, potentially leading to suboptimal analysis. To explore this further we will contrast our results against a conventional Archetypal Analysis model. 

Given the inherent subjectivity of human experience, each individual’s unique perspective can significantly color their interpretation of question items and (Likert) scale points used in the questionnaire, introducing a distinct form of response bias. Response bias presents a significant challenge in cross-national research, as it can distort the validity of survey results and lead to incorrect conclusions \cite{Mazor2002ASurveys}. This phenomenon varies not only across cultures but also across social categories within the same group, reflecting the complexity of human behavior and perception \cite{vanHerk2004ResponseCountries}.

The term response bias has many connotations, two of which we address in this paper. The first instance of response bias is the issue of taking an often qualitative scale and assuming a uniform distance between the possible answers. This is addressed in the proposed Ordinal Archetypal Analysis model (OAA) where the scale during the optimization is optimally changed to a continuous scale which no longer assumes the answers to be equidistant, similar to the first step of \cite{Fernandez2021ArchetypalData}. The second instance is that each subject has its own unique interpretation of scale, reflecting their inherent tendencies. We further introduce the Response Bias Ordinal Archetypal Analysis (RBOAA), which instead of a universal scale, learns individualized scales for each subject during optimization.

Specifically, we derive a direct ordinal AA optimization approach capable of addressing human response biases. Our novel approach extends beyond traditional matrix decomposition methods, offering insights into the intricate fabric of human responses, not only providing new insight into ordinal data but also considering the inherent response bias found in self-reported questionnaire-based data. 
We highlight the procedure on both synthetic as well as a large open source data set, the eight round of the European Social Survey data set from 2016 (ESS8) \cite{Sikt-NorwegianAgencyforSharedServicesinEducationandResearch2016European8}, about personal values across the European Union and Great Britain (GB), where we extract distinct prominent profiles and identify characteristics of different population groups.

\section{METHODS}

Archetypal Analysis was originally proposed by \cite{Cutler1993ARCHETYPALANALYSIS} in the context of least squares minimization defined by
\begin{equation} \label{eq:AAClassic}
\begin{array}{clll}
\min _{\mathbf{C}, \mathbf{S}} & L(\mathbf{X},\mathbf{R})&
\text { s.t. } & \mathbf{R} = \mathbf{XCS} \\
&&&
c_{j, k} \geq 0, \quad s_{k, j} \geq 0 \\
&&& \sum_j c_{j,k} = 1, \quad \sum_k s_{k,j} = 1,
\end{array}
\end{equation}
in which $L(\mathbf{X},\mathbf{R})=\|\mathbf{X}-\mathbf{R}\|^2$, and where $\mathbf{C} \in \mathbb{R}^{N\times K}$ and $\mathbf{S} \in \mathbb{R}^{K\times N}$ are column stochastic matrices. The matrix, $\mathbf{R}$, represents the reconstruction of $\mathbf{X}$ within the imposed constraints.
The AA problem is non-convex in the joint optimization of $\mathbf{C}$ and $\mathbf{S}$, typically solved by alternatingly solving until convergence for $\mathbf{S}$ and $\mathbf{C}$ that respectively form two convex sub-problems \cite{Cutler1993ARCHETYPALANALYSIS,Mrup2012ArchetypalMining}. Whereas the AA framework has been generalized to a variety of other likelihood specifications including binary (Bernoulli) and integer weighted (Poisson) likelihood functions \cite{Seth2016ProbabilisticAnalysis} no principled likelihood based framework has been derived for ordinal variables. Consequently, the existing ordinal AA relies on a two step procedure converting the ordinal scale and subsequently applying conventional least squares AA \cite{Fernandez2021ArchetypalData}.

\begin{figure*}[htbp]
\centering
\subfloat[]{\includegraphics[width=0.32\textwidth]{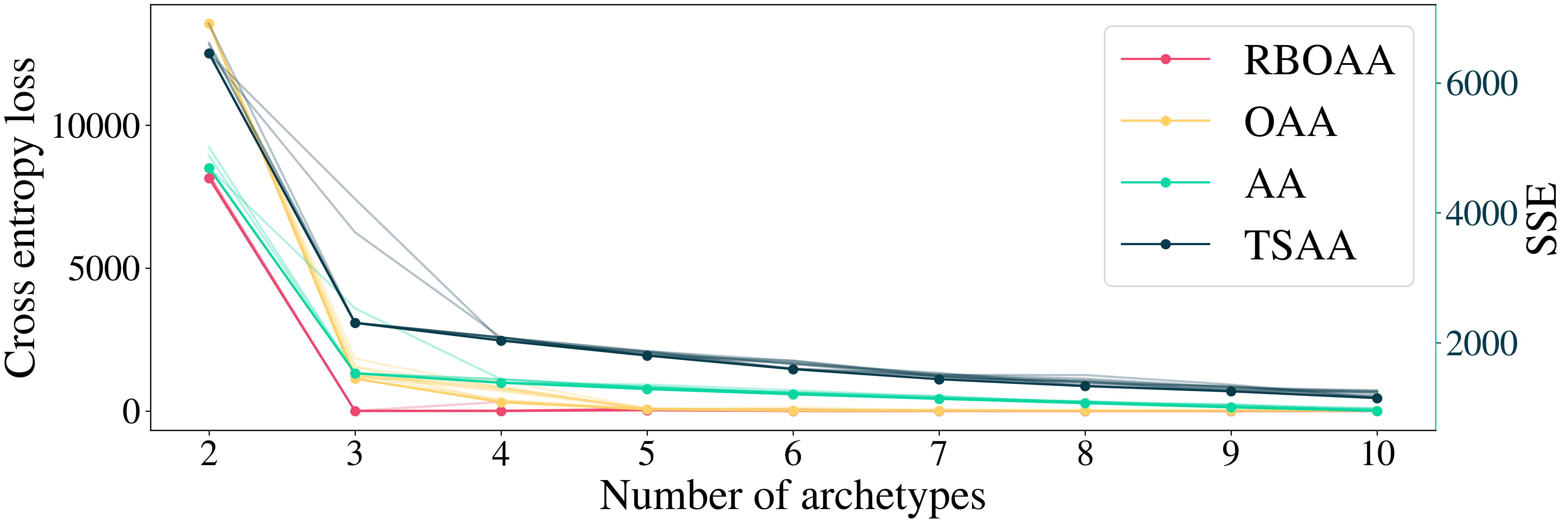}\label{fig:Err_K_NORB}}
\subfloat[]
{\includegraphics[width=0.32\textwidth]{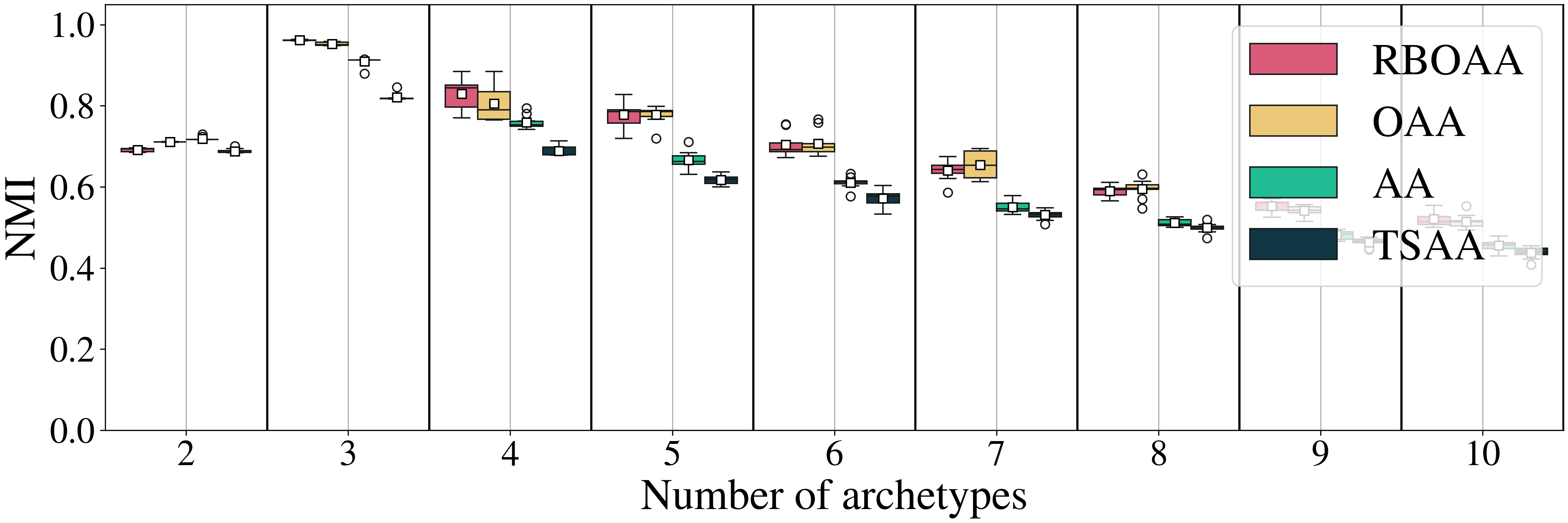}\label{fig:NMI_K_NORB}}
\subfloat[]{\includegraphics[width=0.32\textwidth]{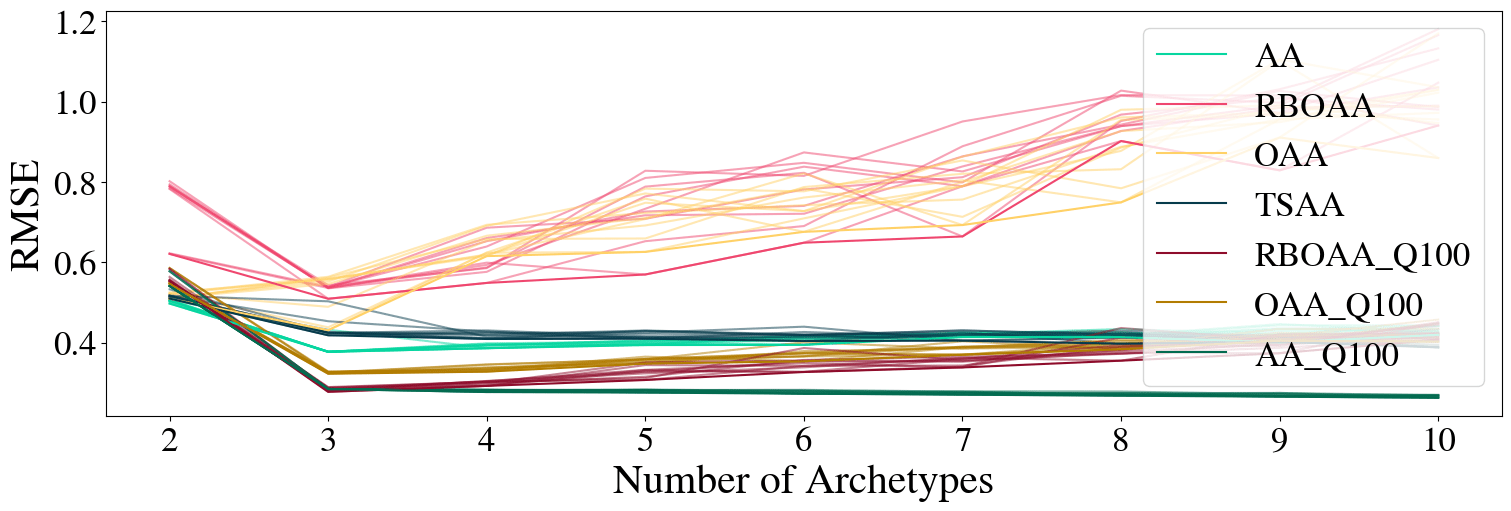}\label{fig:corrError_NORB}}
\vspace{-4mm}
\subfloat[]
{\includegraphics[width=0.32\textwidth]{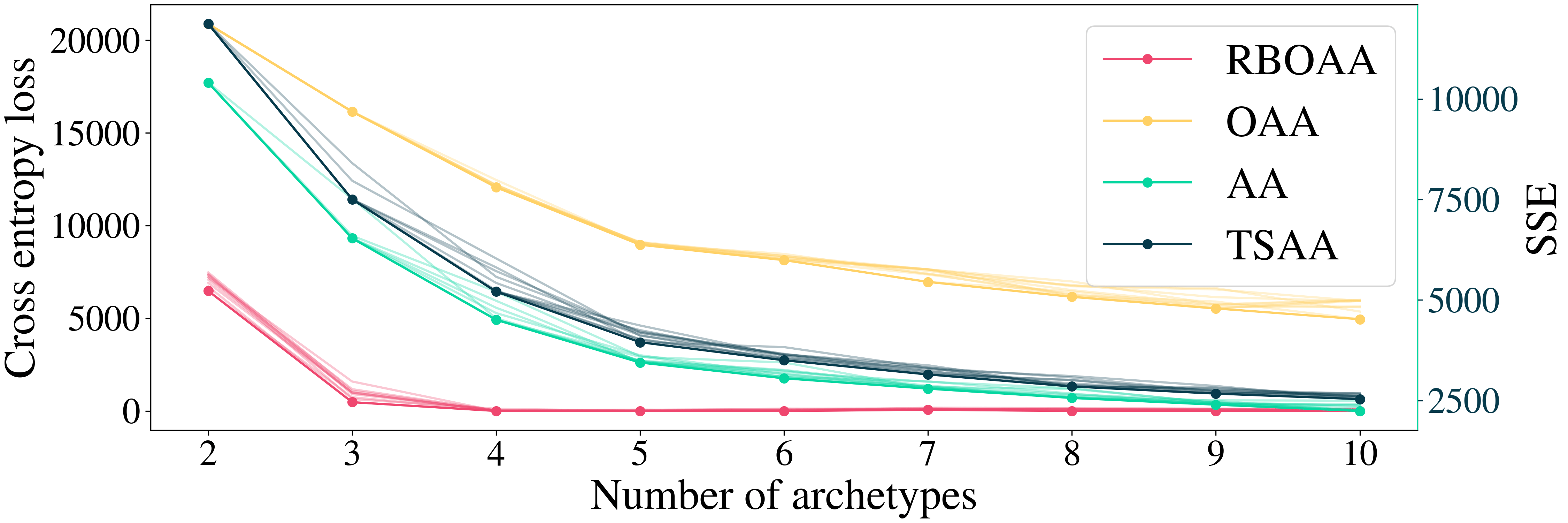}\label{fig:Err_K_RB}}
\subfloat[]{\includegraphics[width=0.32\textwidth]{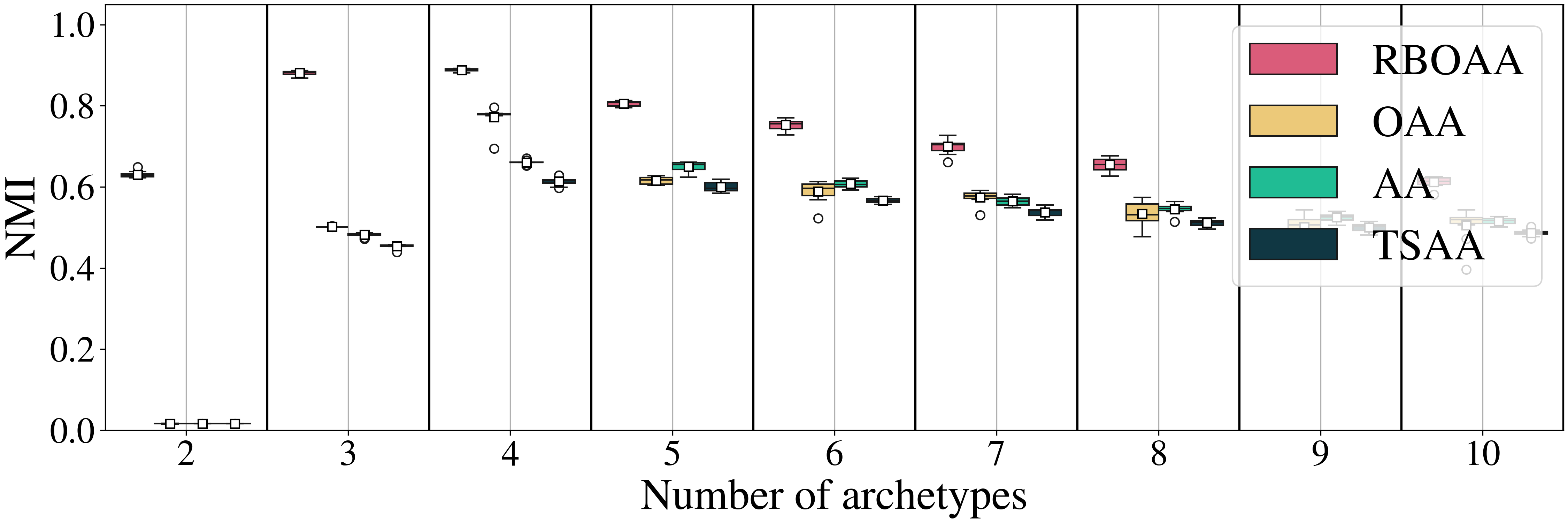}\label{fig:NMI_K_RB}}
\subfloat[]{\includegraphics[width=0.32\textwidth]{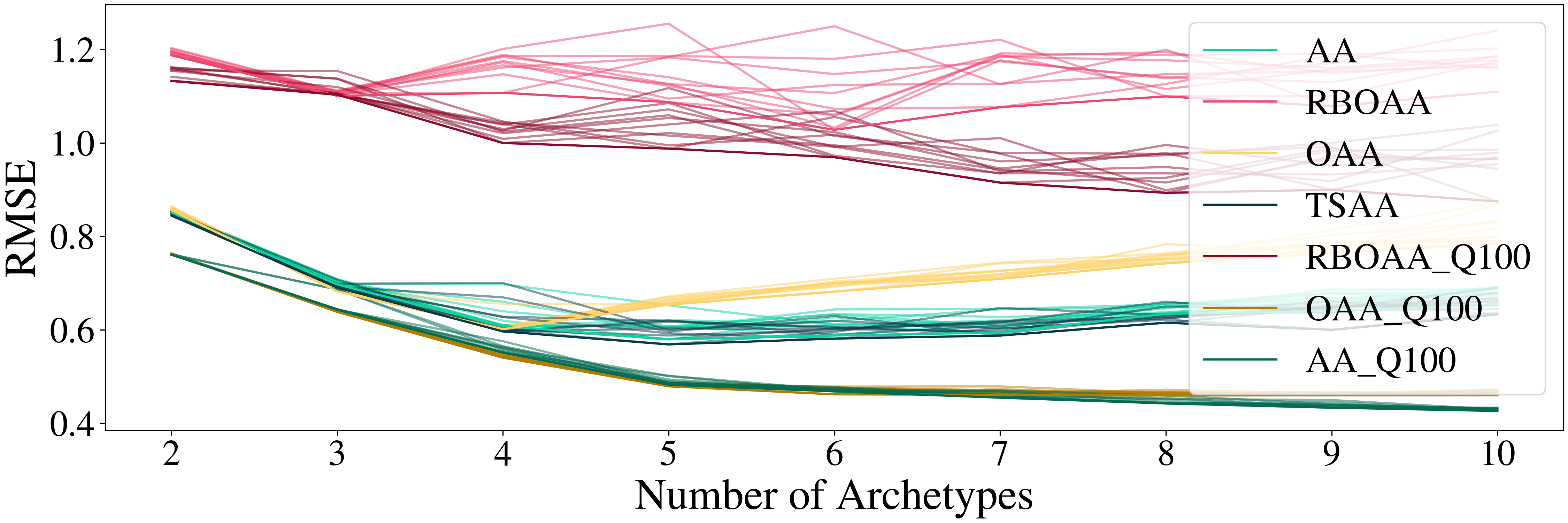}\label{fig:corrError_RB}}
\caption{\textbf{Top panel:} Results of the models on synthetic data without response bias. \textbf{Bottom panel:} Results of the models on synthetic data with response bias. The leftmost column represents the loss in terms of cross-entropy and least-squares across different numbers of archetypes. The middle plots provide the NMI, which in the case of the synthetic data are used as an indication of how well the model reconstructs the ground truth structure in terms of how observations are expressed by the archetypes. The left column is the RMSE between the original data and a corrupted reconstruction, $\mathbf{R}_{cor}$ of the original data.}\label{fig:fig1}
\vspace{-2mm}
\end{figure*}

\subsection{Ordinal Archetypal Analysis}
Given a dataset of $\mathbf{N}$ samples, with ordinal input $x_{m,n} \in \mathbb{Z}$ corresponding to
an answer on a Likert scale with $p$ levels, we assume an unobservable latent function $f(x_{m,n}) \in \mathbb{R}$, which maps an answer to a given question from an ordinal scale to a continuous scale. This is based on defining ordinal boundaries
\begin{equation}
\label{eq:beta_constraints}
 -\infty < \beta_0 <\beta_1< \ldots < \beta_{p} <\infty.
 \end{equation}

In the first model, we disregard response bias, thereby necessitating the computation of an array of boundaries of size $p+1$. Consequently, the model operates under the assumption of uniform response patterns across different respondents. The ordinal likelihood function, initially introduced in \cite{Chu2005GaussianRegression} in the context of Gaussian Process regression, is defined as the probability of observing a specific response given the latent functions. In the absence of noise, this likelihood corresponds to
\begin{equation} \label{eq:noisefree}
  P_{ideal}(x_{m,n}|f(r_{m,n})) = \begin{cases}
  1 & \text{if} ~~ \beta_{x_{m,n}-1} \leq f(r_{m,n}) \leq \beta_{x_{m,n}} \\
  0 & \text{otherwise.}
\end{cases}
\end{equation}

Upon transforming data from an ordinal to a continuous scale, the value of $f(r_{m,n})$ may fall between any two adjacent boundary values. We assume a uniform distribution for the probability of an observation residing within these boundaries. The expected value is thereby given by $\boldsymbol{\alpha}_j$, positioned at the midpoint between its corresponding boundary values, i.e., 
\begin{equation}
\label{eq:alpha}
\boldsymbol{\alpha}_j = \frac{\boldsymbol{\beta}_{j} + \boldsymbol{\beta}_{j-1}}{2}, \quad j \in [1,p],
\end{equation}

The ordinal data matrix $\textbf{X}$ is then mapped to a continuous domain $\tilde{\textbf{X}}$ through (\ref{eq:alpha}). This transformation replaces all elements of $\textbf{X}$ with their corresponding $\boldsymbol{\alpha}$-values, as determined by the likelihood function. Subsequently, archetypal analysis is applied to $\tilde{\textbf{X}}$, resulting in
\begin{equation}
  \mathbf{R} = \tilde{\textbf{X}}\textbf{C}\textbf{S}.
\end{equation}

In the typical scenarios where noise is present in the ordinal observations of $\textbf{X}$, the latent functions $f(r_{m,n})$ are accompanied by Gaussian noise $\delta_{m,n} \sim \mathcal{N}(0, \sigma^2)$. This introduces a modified likelihood function \cite{Chu2005GaussianRegression}
\begin{equation}\label{eq:likelihoodFinal}
    \begin{aligned}
    P(x_{m,n}|f(r_{m,n})) &= \int P_{ideal}(x_{m,n}|f(r_{m,n}) + \delta_{m,n}) \\
    &\cdot \mathcal{N}(\delta_{m,n};0, \sigma^2) \, \text{d}\delta_{m,n} \\
    &= \Phi(z_{x_{m,n}}^{m,n}) - \Phi(z_{x_{m,n}-1}^{m,n}),
    \end{aligned}
\end{equation}
where $z_{x_{m,n}}^{m,n}$ and $z_{x_{m,n}-1}^{m,n}$ represent the standardized distances from an observation to the upper and lower boundaries, respectively, and $\Phi(z)$ denotes the cumulative distribution function of the standard normal distribution obtained from the convolution of a uniform interval with the Gaussian noise-kernel, such that

\begin{equation}
    \begin{aligned}
        z^{m,n}_{j} &= \frac{\beta_j - r_{m,n}}{\sigma}, \quad 
        z^{m,n}_{j-1} = \frac{\beta_{j-1} - r_{m,n}}{\sigma}, \\
        \Phi(z) &= \int_{-\infty}^z \mathcal{N}(z;0,1) \, dz,
    \end{aligned}
\end{equation}
thereby, the likelihood function can be conceptualized as fitting a Gaussian kernel over the boundaries, with the parameter $\sigma$ accounting for the potential leakage of an observation in $\textbf{R}$ to adjacent boundaries.

\subsection{Response Bias Ordinal Archetypal Analysis}
The OAA model assumes consistent perceptions of scale across individuals, which is not always a valid assumption. The ordinal AA method can be advanced to account for individual perceptions of the ordinal scale by defining subject specific boundaries that for the $n^{th}$ subject is given by $\boldsymbol{\beta}_n$. This yields N sets of boundary vectors ${\boldsymbol{\beta}_1,\dots, \boldsymbol{\beta}_N}$ each with $p+1$ elements.
\begin{equation}
    -\infty < \beta_{n,0} < \beta_{n,1}< \ldots < \beta_{n, p} < \infty,
\end{equation}
Consequently, the sets of $\boldsymbol{\alpha}$-values are computed from the individualized boundaries. Hence, there is correspondingly a total of $N \times p$ values in $\boldsymbol{\alpha}$ given by:
\begin{equation}\label{eq:alphaRBOAA}
    \alpha_{n,j} = \dfrac{\beta_{n,j} + \beta_{n,j-1}}{2},
\end{equation}
furthermore, the $sigma$ parameter is similarly computed for each individual i.e.  $\boldsymbol{\sigma} \in \mathbb{R}^\text{N}$ defining different subject specific noise levels in the response patterns. As a result, the uncertainty wrt. an individual's response is considered rather than accounting only for the overall uncertainty across all respondents.

\subsection{Model Optimization}
The implementation of our method is conducted in Python, utilizing the robust modeling framework provided by PyTorch. We optimize the likelihood function, as delineated in (\ref{eq:likelihoodFinal}), employing the AMSGrad variant of the Adam optimizer. This optimization is conducted with a learning rate of 0.1 for AA and 0.01 for OAA and RBOAA, respectively, complemented by an early stopping mechanism. The OAA and RBOAA models are initialized with a few iterations of conventional AA to reduce local minima. We implement the AA sum to one and non-negativity constraints on $\mathbf{C}$ and $\mathbf{S}$ using the softmax function thereby reparameterising the optimization of $\mathbf{C}$ and $\mathbf{S}$ in terms of unconstrained variables $\mathbf{\tilde{C}}$ and $\mathbf{\tilde{S}}$ such that $\mathbf{c}_k=softmax(\mathbf{\tilde{c}}_k)$ and $\mathbf{s}_n=softmax(\mathbf{\tilde{s}}_n)$. 

For the OAA and RBOAA we repameterize $\sigma$ using the softplus function $\sigma=\log(1+\exp(\tilde{\sigma}))$. Furthermore, the boundary values $\boldsymbol{\beta}$ are optimized by reparameterizing 

\begin{equation}\label{eq:betas}
   \boldsymbol{\beta} = \log(1+\exp(c_1)) \cdot \mathbf{b} + c_2,
\end{equation}
such that \begin{equation}
\label{eq:b_constraints}
    0 = b_0<b_1< \ldots < b_{p-1}<b_p = 1,
\end{equation}

Using this reparametierization we enforce the monotonicity constraints on the values of $\boldsymbol{b}$ using a softmax function in combination with a cumulative sum, i.e.
\begin{equation}
b_0=0,\quad 
    b_l = \sum_{j=1}^{l} \frac{e^{\tilde{b}_j}}{\sum_{i=1}^{p}e^{\tilde{b}_i}},
\end{equation}
thereby enforcing the constraints given in (\ref{eq:beta_constraints}). For the RBOAA model we similarly specify subject specific parameters $\boldsymbol{b}_n$, $c_{n,1}$ and $c_{n,2}$ for the optimization of $\boldsymbol{\beta}_n$.

We initialize $\mathbf{S}$ and $\mathbf{C}$ by randomly sampling from a uniform Dirichlet distribution.
The $\Tilde{b}_j$ are initialized as $\boldsymbol{1}$. This ensures equidistant boundaries when constraints are applied prior to applying the optimization framework.
Furthermore, $\tilde{\sigma}$ was initialized using the standard normal distribution. In the rare instances of numerical issues of the initialization in which the loss function returns NaN we re-initialized the model. The code is publicly available on \href{https://github.com/Maplewarrior/OrdinalArchetypalAnalysis?}{Github}.


\subsection{Model Evaluation}
The RBOAA models were optimized in terms of the cross-entropy loss between the estimated probability of the true ordinal category whereas the two-step Archetypal Analysis (TSAA) procedure where the data is converted to continuous scale\cite{Fernandez2021ArchetypalData}
 and conventional AA both used least-squares minimization, corresponding to optimizing a standard Gaussian likelihood.

We further evaluated the models in their ability to reconstruct corrupted data to assess how prone the models were to overfitting. For this assessment we trained the model on the corrupted data to predict the entries prior to corruption. To make the assessment comparable across all the models; AA, TSAA, OAA, and RBOAA, we used the root-mean-square error (RMSE) as performance metric. As the TSAA model converts the scale before performing conventional AA we find the reconstruction $\mathbf{R}$ based on the original data prior to converting the scale, i.e. corresponding to the prediction from a conventional AA, to ensure comparisons were performed on the same scale. As the OAA and RBOAA are based on cross-entropy minimization and returns the probability for each ordinal value we converted this to the expected predicted response $\hat{r}_{n,m}$ using
\begin{equation}
E(\hat{r}_{n,m})=\sum_j j \cdot P(j|f(r_{n,m})),
\end{equation}

Finally, we also quantify the consistency of the the observations characterizations in terms of archetypes as defined through $\mathbf{S}$ by quantifying the normalized mutual information (NMI) as proposed in \cite{Hinrich2016ArchetypalData}.

\section{Results and discussion}
\subsection{Synthetic study}

To demonstrate the proposed model frameworks, we evaluate them on two synthetic data sets respectively designed with and without response bias according to the OAA and RBOAA model formulations.
The data is generated to have three archetypes, 20 questions and 1000 respondents with an ordinal scale of 5. We compare the two ordinal models (OAA and RBOAA) to the two existing models, namely the original Archetypal Analysis model (AA)\cite{Cutler1993ARCHETYPALANALYSIS}, where the ordinal data is treated as continuous and the TSAA of \cite{Fernandez2021ArchetypalData}. Both optimization problems are solved using conventional least-squares minimization.
\begin{figure}%
  \centering
    \subfloat[Visualization of the models response bias for data with no response bias]{{\includegraphics[width=0.4\linewidth]{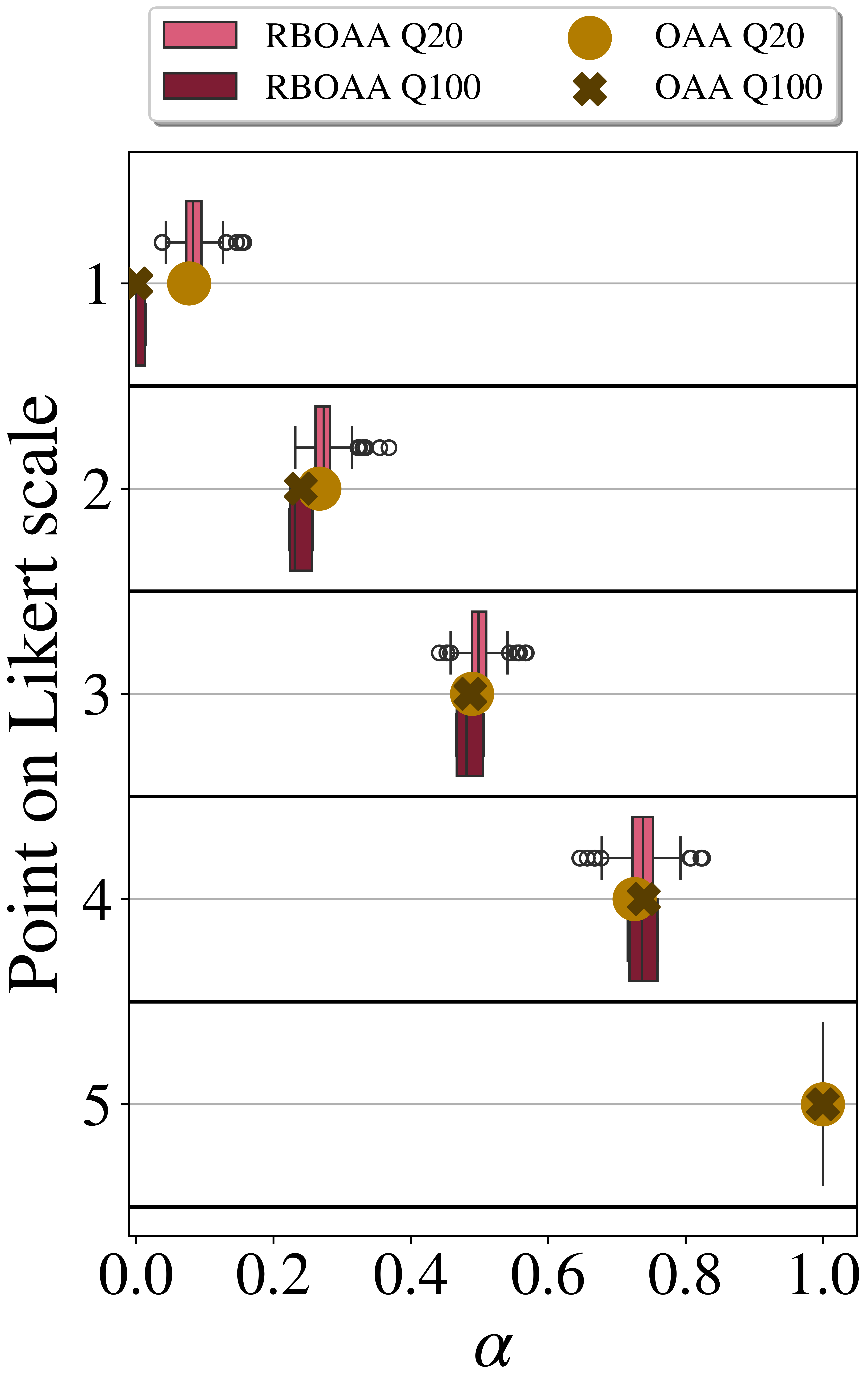} }\label{fig:No_RB} }%
    \qquad
    \subfloat[Visualization of the models response bias for data with response bias]{{\includegraphics[width=.4\linewidth]{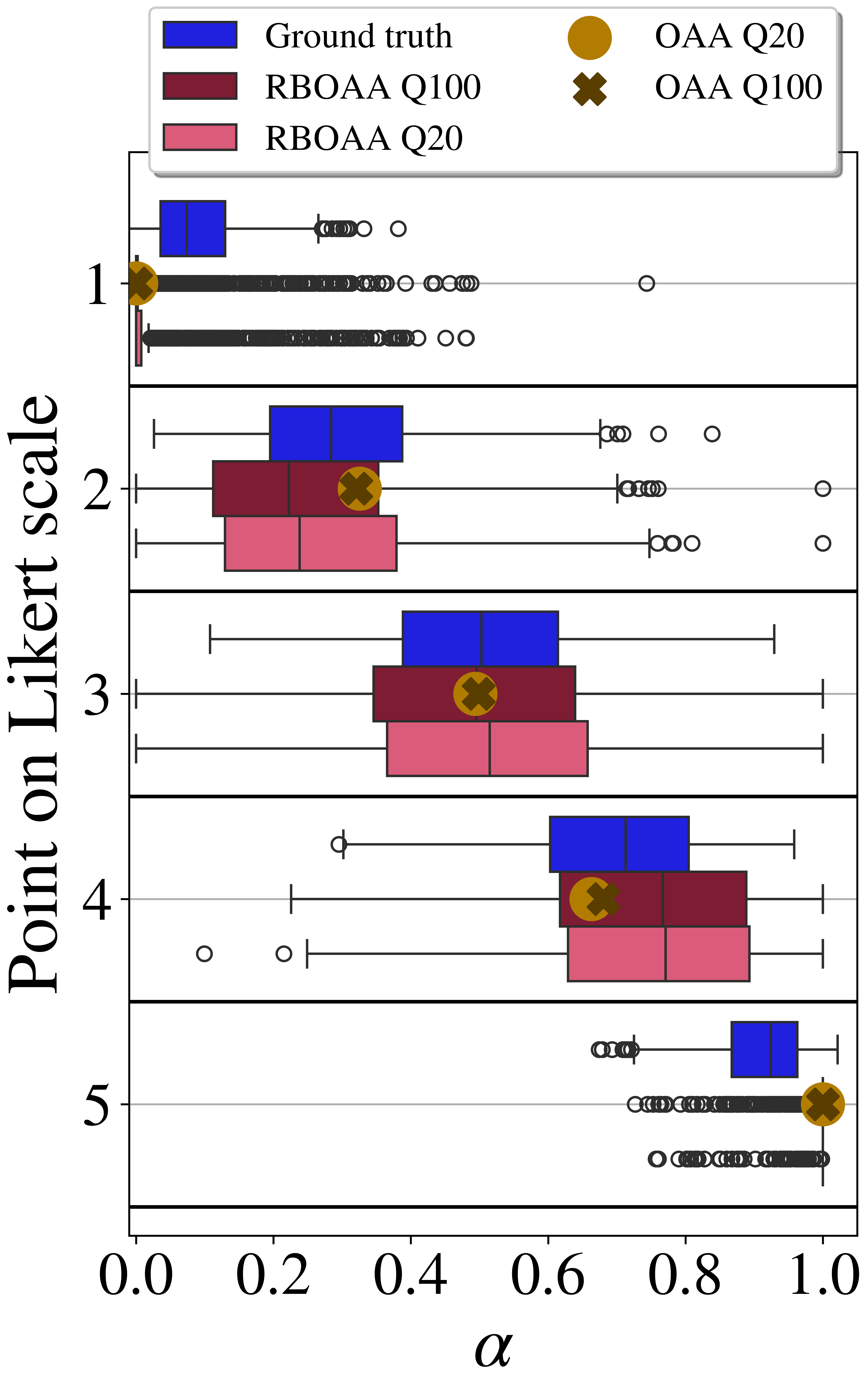} } \label{fig:RB}}%
    \caption{Response bias found the the OAA (points) and RBOAA (boxplot) for the synthetic data without response bias (a) and with response bias (b) together with the ground truth. In the case with no response bias the ground truth are evenly spaced between 0 and 1.a}%
    \label{fig:SynthRB}%
\end{figure}
In fig.~\ref{fig:fig1}, a comparative analysis of the AA, TSAA, OAA, and RBOAA are given. Whereas OAA and RBOAA optimize cross-entropy loss, TSAA and AA are optimized using the least-squares loss, making direct comparisons impossible outside of archetype function improvement. Notably, all models showed comparable performance on data with no response bias, but OAA and RBOAA have a distinct advantage in identifying the true structure of observations as expressed by archetypes (i.e., $NMI(\mathbf{S}^{\text{estimated}},\mathbf{S}^{\text{true}}$). The RBOAA model, in particular, demonstrated superior performance in datasets with response bias, effectively indicating a three-archetype solution without significant loss improvements beyond this point. This solution also showed a high Normalized Mutual Information (NMI) correspondence with the original data's archetype expression. However, we also observe that the response bias modeling comes with a price of overfitting to noisy observations as we see the reconstruction of the uncorrupted entries when training on corrupted data performs inferior to the modeling approaches not including response bias modeling, i.e. OAA, AA and TSAA that perform similarly. To examine this further two larger synthetic datasets were created with 100 questions instead of 20, but otherwise initialized with the same hyper-parameters. For the instance with no response bias, we observe that the denoising clearly improved upon the RBOAA and OAA models' performance. For the model with response bias, the RBOAA model still overfits to the data. We do observe improved performance, but an even larger synthetic data set is needed to examine this further (fig.~\ref{fig:corrError_NORB} and \ref{fig:corrError_RB}). 

In fig.~\ref{fig:SynthRB} the learned response biases are compared to the ground truth perceptions of scale used for the synthetic datasets with and without response bias. We generally observe a high degree of overlap for RBOAA in both settings to the ground truth values. The OAA model performs slightly worse on the dataset with response bias. This aligns with the notion that treating scales as constant across respondents in the presence of response bias is suboptimal.

\subsection{ESS Round 8 (2016) data}
The European Social Survey (ESS) is a cross-national survey conducted biennially across Europe since 2001. It involves face-to-face interviews with cross-sectional samples, measuring attitudes, beliefs, and behavior patterns in over thirty nations. We consider a subpopulation of the data, namely Great Britain, to allow for comparison with the TSAA model that does not scale well to the full data. This subpopulation consists of 1897 responders, answering 21 questions, measuring ten basic human values\cite{Schwartz2012AnValues}, in this paper four of these values have been highlighted. 

\begin{figure}
    \centering
    \begin{tabular}{cc}
    \adjustbox{valign=b}{\subfloat[Response bias\label{fig:RWD_RB}]{%
          \includegraphics[width=.25\linewidth]{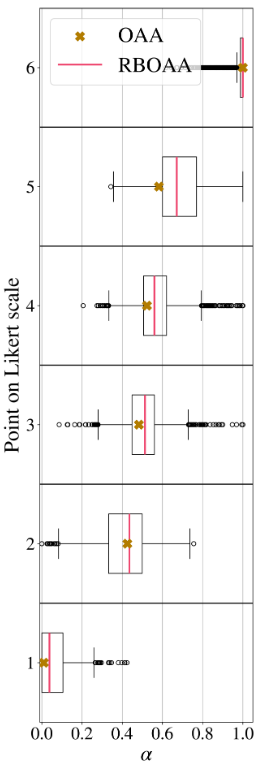}}}
          \vspace{-3mm}
    &      
    \adjustbox{valign=b}{\begin{tabular}{@{}c@{}c@{}}
    \subfloat[Error\label{fig:GB_loss}]{%
          \includegraphics[width=.6\linewidth]{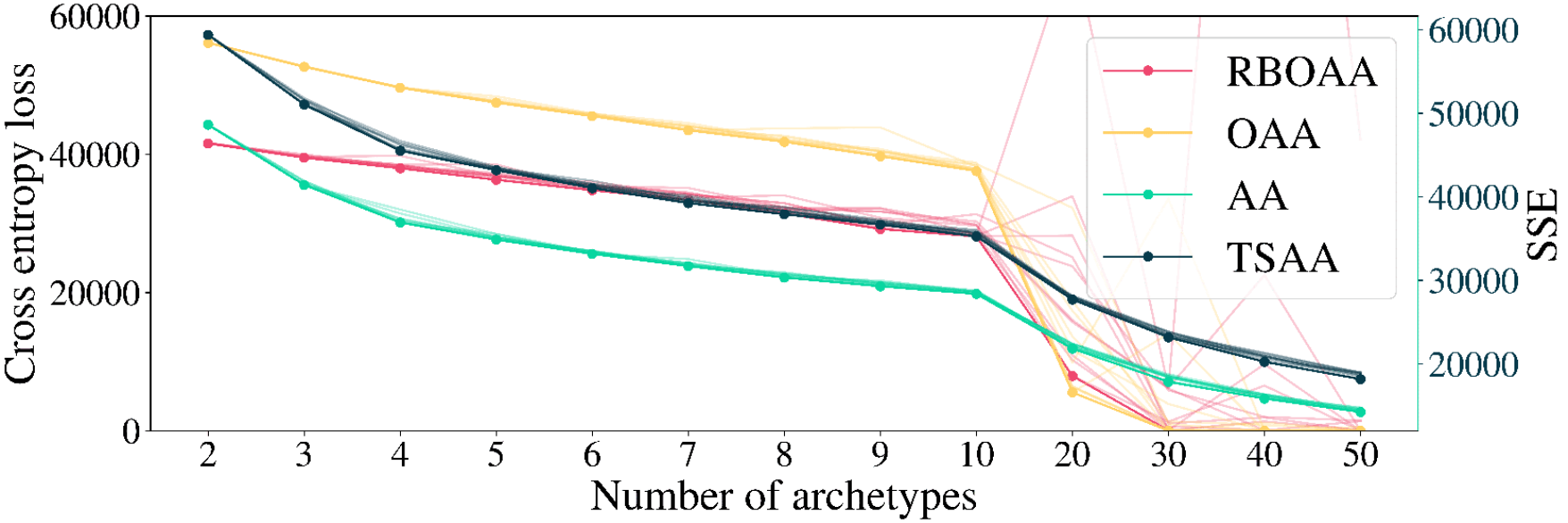}}\\
          \vspace{-3mm}
    \subfloat[NMI\label{fig:GB_NMI}]{%
          \includegraphics[width=.6\linewidth]{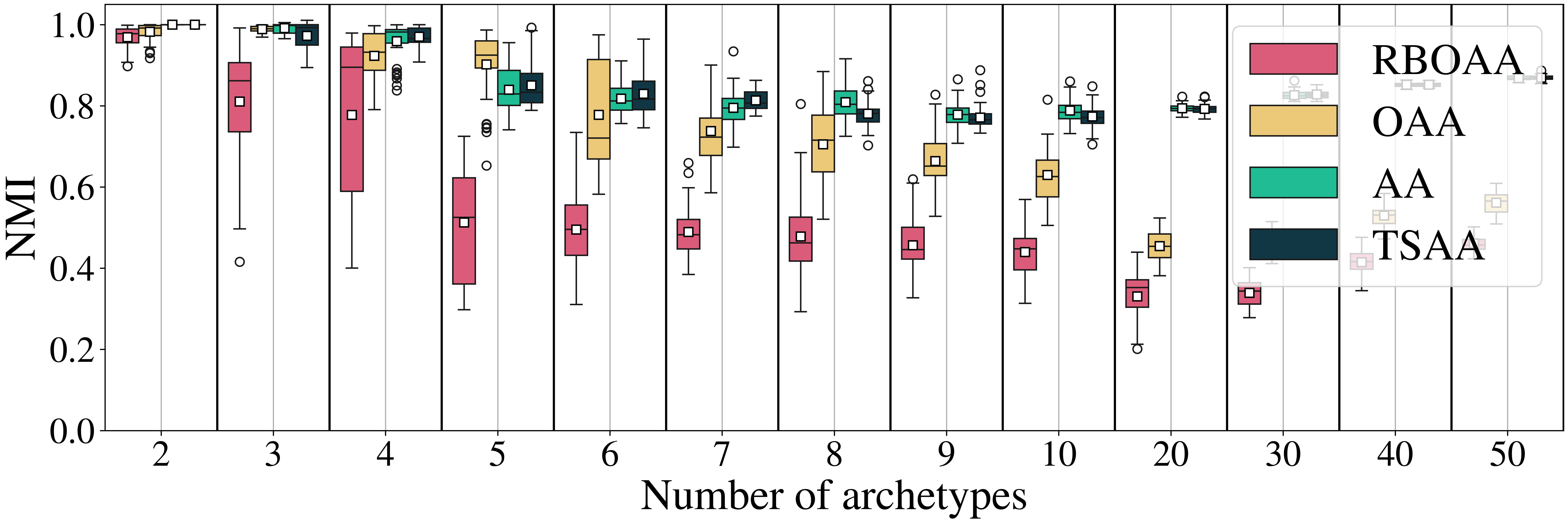}}\\
    \subfloat[Corruption Error\label{fig:GB_cor}]{%
          \includegraphics[width=.6\linewidth]{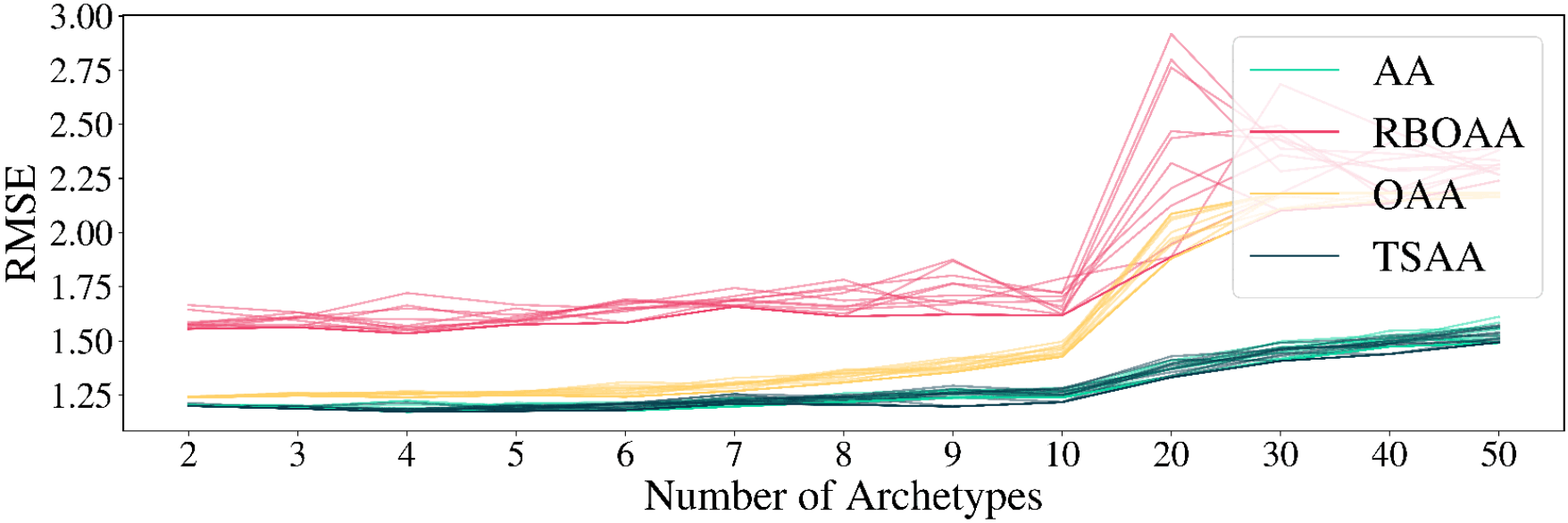}}
    \end{tabular}}
    \end{tabular}
    \caption{The loss and NMI across a different number of archetypes for ESS8 GB data as well as the response bias }
    \vspace{-1mm}
    \label{fig:GB_eval}
  \end{figure}

From fig.~\ref{fig:GB_eval}, it is evident that both the OAA and RBOAA models encounter significant local minimas, when initialized with a higher number of components. In contrast, AA and TSAA models exhibit remarkable stability, even as the number of components increases. Importantly, the ability of OAA and RBOAA to learn the ordinal scale directly in a single step is a significant advantage, offering a streamlined process that can be extended to ascertain the individual ordinal scales of each subject by RBOAA. In fig.~\ref{fig:RWD_RB} the response bias across all subject has been visualized as boxplots for each answer corresponding to points on a Likert scale and we here observe that the RBOAA learns substantial variability in terms of the respondents perception of scale. However, the observed overfitting in fig.~\ref{fig:corrError_RB} underscores the need for careful model evaluation and possibly the incorporation of regularization techniques in the RBOAA to prevent the model from fitting noise.

\begin{figure*}[htbp]

\centering
\subfloat[]
{\includegraphics[height = 6.8cm]{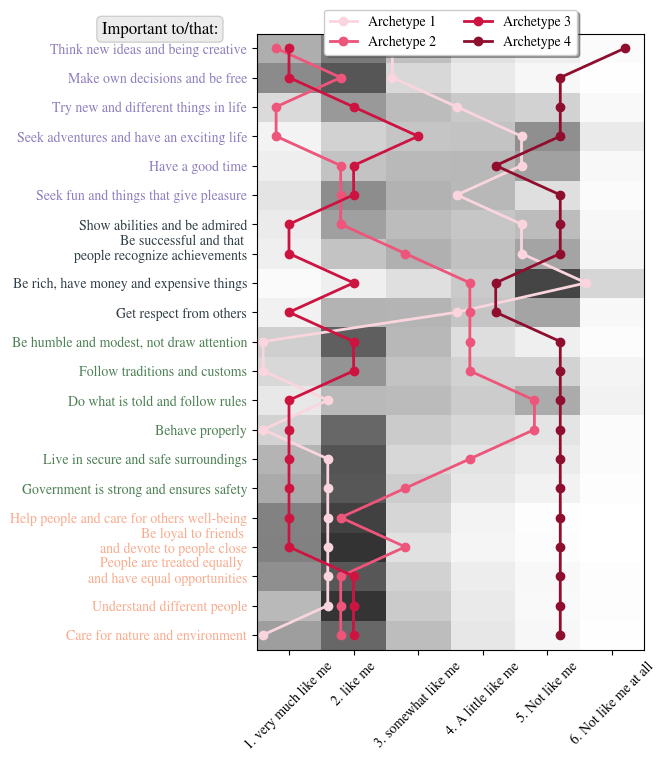}\label{fig:Arc_OAA}} \hspace*{-1mm}
\subfloat[]{\includegraphics[height = 6.8cm]{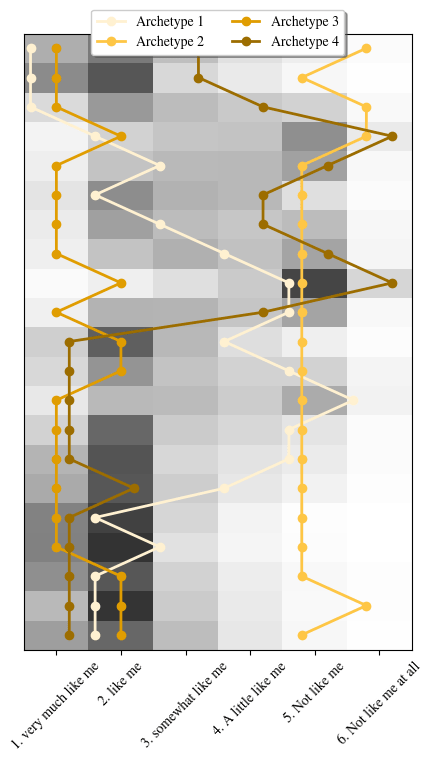}
\label{fig:Arc_RBOAA}} \hspace*{-2mm}
\subfloat[]
{\includegraphics[height = 6.8cm]{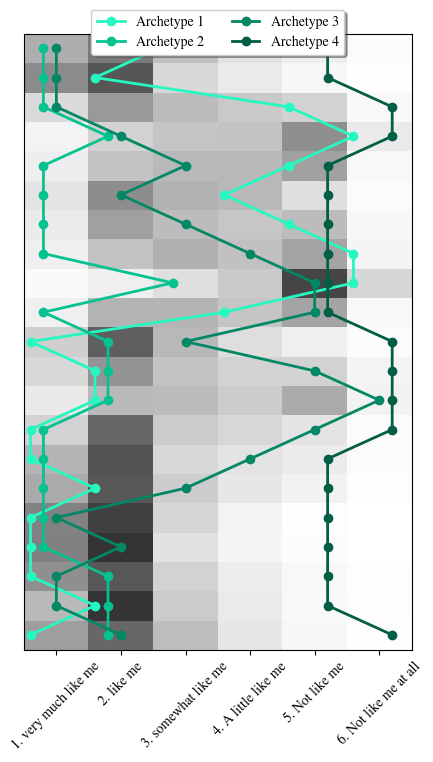}\label{fig:Arc_AA}} \hspace*{-1mm}
\subfloat[]{\includegraphics[height = 6.8cm]{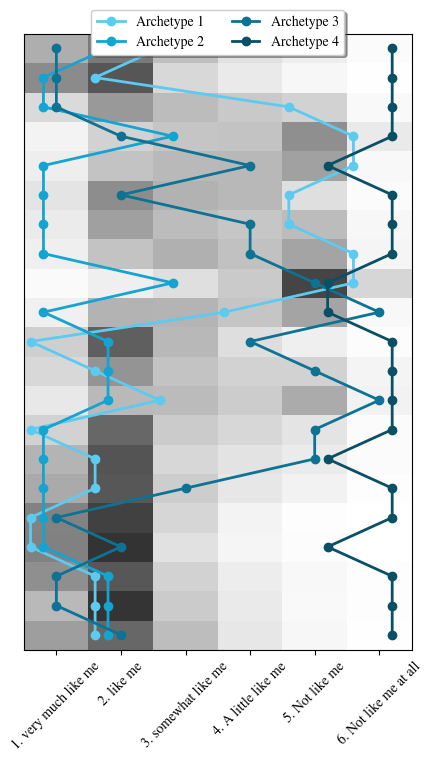}\label{fig:Arc_TSAA}}
\caption{From left to right: The four archetypes for each of the models; (a) RBOAA, (b) OAA, (c) AA and (d) TSAA. The background is colored after the subjects answer to each question, a darker color indicating that the answer is popular among the subjects. The different archetypes have been visualized for each model as trajectories (vertical lines) across the questions, highlighting their distinct profile. The shared labels of the y-axis have been colored after the questions category according to Schwartz theory of human values \cite{Schwartz2012AnValues}, \textcolor{lavendar}{\textbf{Openness to change}}, \textcolor{pine}{\textbf{Self Enhancement}}, \textcolor{grassgreen}{\textbf{Conservation}} and \textcolor{peach}{\textbf{Self Transcendence}} }\label{fig:RWDArchetypes}
\end{figure*}

Determining the optimal number of archetypes on the ESS8 data is a more challenging task. There is no clear improvement on the loss curve for a given number of archetypes (fig.~\ref{fig:GB_loss}). To enhance the information in the model we therefore chose the maximum number of components that the model can have, without suffering from loss of stability (NMI). From fig.~\ref{fig:GB_NMI} $K=4$ is chosen. Based on this we extract the archetypes from the repetition resulting in the lowest error, fig.~\ref{fig:Arc_OAA}-\ref{fig:Arc_TSAA}. From these figures, the trajectory of each archetype through the questions can be seen. It is clear that we observe similar patterns in the archetypes across the models, suggesting both an archetype that identifies heavily with the questions and one that does not. From the questions there are also indications of a more conservative and a rebellious archetype.

\section{Conclusion}
We have proposed the ordinal archetypal analysis directly providing a principled procedure for the modeling of ordinal data by AA. Notably, we extended the approach to account for response bias and demonstrated on synthetic data how this improved recovery of how observations were characterized in terms of the archetypes in the presence of biased responses. However, we also observed that the added flexibility when facing a limited number of questions (i.e., 20 questions pr. respondent) resulted in overfitting and the inability to well recover noisy observations when including the response bias modeling. The proposed approach readily scales to large datasets and future work should explore how the methodology can improve upon the understanding of human responses by characterizing distinct response patters as enabled by the proposed ordinal archetypal analysis framework.

\bibliographystyle{IEEEtran}
\bibliography{references}

\end{document}